\documentclass{article}

\usepackage{arxiv}

\usepackage[utf8]{inputenc} % allow utf-8 input
\usepackage[T1]{fontenc}  
\usepackage[linesnumbered,ruled,vlined]{algorithm2e} % use algorithm2e with numbered lines, ruled and vlined style
\usepackage{hyperref}       % hyperlinks
\usepackage{url}            % simple URL typesetting
\usepackage{booktabs}       % professional-quality tables
\usepackage{amsfonts}       % blackboard math symbols
\usepackage{nicefrac}       % compact symbols for 1/2, etc.
\usepackage{microtype}      % microtypography
\usepackage{lipsum}         % Can be removed after putting your text content
\usepackage{graphicx}
\usepackage{natbib}
\usepackage{multirow}
\usepackage{doi}
\usepackage{amsmath,amssymb,amsfonts}
\usepackage{subcaption} % Load the subcaption package
\usepackage{comment}

\title{Fast Clustering of Categorical Big Data}

\author{\thanks{Corresponding Author: Bipana Thapaliya, bipana.thapaliya03@gmail.com.}\, Bipana Thapaliya \textsuperscript{1} \quad \textbf{Yu Zhuang \textsuperscript{1}}\\
\textsuperscript{1}Texas Tech University  \quad 
\\}

\renewcommand{\shorttitle}{\textit{Fast Clustering of Categorical Big Data} }

\begin{document}
\maketitle

\begin{abstract}
The K-Modes algorithm, developed for clustering categorical data, is of high algorithmic
simplicity but suffers from unreliable performances in clustering quality and clustering
efficiency, both heavily influenced by the choice of initial cluster centers. In this
paper, we investigate Bisecting K-Modes (BK-Modes), a successive bisecting process to 
find clusters, in examining how good the cluster centers out of the bisecting process 
will be when used as initial centers for the K-Modes. The BK-Modes works by splitting 
a dataset into multiple clusters iteratively with one cluster being chosen and bisected 
into two clusters in each iteration.  We use the sum of distances of data to their 
cluster centers as the selection metric to choose a cluster to be bisected in each 
iteration. This iterative process stops when $K$ clusters are produced. The centers 
of these $K$ clusters are then used as the initial cluster centers for the 
K-Modes. Experimental studies of the BK-Modes were carried out and were compared 
against the K-Modes with multiple sets of initial cluster centers as well as the 
best of the existing methods we found so far in our survey.  Experimental results 
indicated good performances of BK-Modes both in the clustering quality and efficiency 
for large datasets.
\end{abstract}

\section{Introduction}
Clustering is frequently used in data mining and analysis applications across various domains. Recently, beyond its traditional applications, clustering techniques has been effectively integrated with deep learning techniques \cite{CLuster_1, Cluster_2, Thapaliya_ECGN}, broadening its impact by enabling enhanced data representation and decision-making in fields such as social network analysis \cite{Mishra, deep_clustering} and healthcare \cite{Thapaliya2024, Thapaliya2025, Thapaliya_DSAM, SCGT}.
A clustering process is to group data into clusters with high similarity for data within each cluster but large dissimilarity between different clusters. The K-Means algorithm \cite{macqueen1967some} \cite{anderberg1973cluster} and its variants are probably the most successful clustering methods \cite{macqueen1967some,bai2013fast}. Utilizing the property that the mean minimizes the sum of square distance in the Euclidean distance, the K-Means algorithm is best suitable for numerical data with the Euclidean distance as the data dissimilarity metric. 

For categorical data, Euclidean distance is no longer meaningful, and the Hamming distance is a natural choice. For a categorical data set with each data point consisting of $m$ attributes (or dimensions), the Hamming distance $d(x,y)$ between two data points 
\[
x=\{x_1,x_2,\ldots,x_m\} \quad \text{and} \quad y=\{y_1,y_2,\ldots,y_m\}
\]
is defined by the number of mismatches for all attributes, that is,
\begin{equation}
d(x, y) = \displaystyle\sum_{i=1}^{m} \delta(x_i , y_i),
\end{equation}
where 
\begin{equation}
\delta(x_i , y_i) = \begin{cases} 0 & (x_i = y_i) \\ 1 & (x_i \neq y_i) \end{cases}.
\end{equation}
When the Hamming distance is used as the data dissimilarity measure, it is known \cite{huang1997fast} that the mode of a dataset $S$, the data point whose value of each attribute is the most frequent value for that attribute, minimizes the sum of dissimilarities, that is,
\begin{equation}
\displaystyle {\rm Argmin}_{x \in S} \sum_{s \in S} d(s, x).
\end{equation}
Utilizing this property of the mode in minimizing the sum of distances, the K-Mode algorithm \cite{huang1997fast,huang1998extensions} was developed for categorical data with a simplistic algorithmic procedure the same as that of the K-Means but with the means replaced by the modes. The K-Modes differs from the K-Means algorithm in the following aspects:
\begin{itemize}
\item K-Means uses the Euclidean distance while K-Modes uses the Hamming distance.
\item K-Means uses the mean as a cluster center while K-Modes uses the mode.
\end{itemize}
With a given value of $K$, the K-Modes algorithm consists of the following steps:
\begin{enumerate}
\item[1.] Select $K$ initial cluster centers, one for each cluster;
\item[2.] For each data point, compute its distance to the $K$ cluster centers, and assign the data point to the cluster whose center is closest;
\item[3.] Update the centers of all $K$ clusters by computing the mode of each cluster.
\item[4.] Repeat steps 2--3 until no data point changes the cluster it is assigned.
\end{enumerate}

K-Modes possesses the same algorithmic simplicity as K-Means, as well as a shared weakness: its performance is significantly influenced by the choice of the initial $K$ cluster centers. Huang \cite{huang1997fast,huang1998extensions} proposed two ways for choosing the initial $K$ centers (one by selecting the first $K$ data points, and the other by assigning the most frequent category value of each attribute to the centers). However, these methods were found to be unreliable. In this paper, we target large datasets—where efficiency is as important as clustering quality—and propose a method for choosing initial centers that is both highly efficient and produces high clustering quality.

The rest of the paper is organized as follows. In Sec. 2 we provide a survey of existing work on selecting initial cluster centers for K-Modes. In Sec. 3 the proposed new initialization method is presented. In Sec. 4 experimental tests, the clustering evaluation metric, and the experimental results are discussed. We summarize our work in Sec. 5.

\section{Survey of Related Work} %%%%%%%%%%%%%%%%%% Section 2 %%%%%%%%%%%%%%%%%%
We first list commonly used notations in this and later sections. We use $K$ to denote the number of clusters into which a dataset is to be partitioned, $X$ to denote a categorical dataset with $n$ data points, and $M$ to denote the set of attributes (with $m$ attributes). The notation $d(\cdot,\cdot)$ is used to denote the Hamming distance between two data points.

\subsection{Sun, Zhu and Chen, 2002}\label{sec: sun_zhu_2002}
Sun, Zhu, and Chen \cite{sun2002iterative} proposed an initial center selection method based on Bradley and Fayyad's algorithm \cite{bradley1998refining}. The idea is to draw $J$ small random subsets $S_i$ from $X$, apply K-Modes clustering (using random initialization) to each subset (yielding $J$ sets of solutions $CS_i$, each containing $K$ centers), and then take the union $CS$ of all these centers. Next, $CS$ is clustered $J$ times to produce refined centers, and the candidate with the minimum distortion is selected as the initial set for K-Modes.

\subsection{Khan and Ahmad, 2003}
Khan and Ahmad \cite{khan2003computing} proposed an initial center selection algorithm based on a density-based multiscale data condensation approach \cite{mitra2002density}, replacing the Euclidean distance with the Hamming distance. The algorithm is summarized in Algorithm~\ref{algo:khanAndAhmad2003}.

\begin{algorithm}
\DontPrintSemicolon
\KwIn{$X$, $K$}
\KwOut{Set $C$ containing $K$ initial centers}
$C \gets \emptyset$\;
$k \gets 2$\;
\While{number of data points in $C < K$}{
   \ForEach{$x \in X$}{
    Calculate the distance from $x$ to its $k$-th nearest neighbor, denoted by $r_{k,x}$.
   }
   Select the point in $X$ having the lowest value of $r_{k,x}$ and add it to $C$.\;
   Remove all points from $X$ that lie within a disc of radius $2r_{k,x}$ centered at that point.\;
   $k \gets k + 1$\;
}
\Return{$C$}\;
\caption{K-Modes initialization based on density-based multiscale data condensation \cite{khan2003computing}.}
\label{algo:khanAndAhmad2003}
\end{algorithm}

\subsection{He, 2006}
He \cite{he2006farthest} proposed an initial center selection algorithm for K-Modes using a farthest-point clustering heuristic \cite{gonzalez1985clustering}. The method selects the first center (either randomly or by a scoring function that sums the frequencies of attribute values) and then iteratively selects the point that maximizes the minimum distance to the already chosen centers.

\subsection{Khan and Kant, 2007}
Khan and Kant \cite{khan2007computation} proposed an initial center selection method based on evidence accumulation. The K-Modes algorithm is run $N$ times with random initialization, and the resulting $K$ modes from each run are pooled. The most frequently occurring modes are then selected as the initial centers. This process is summarized in Algorithm~\ref{algo:khanAndKant2007}.

\begin{algorithm}
\DontPrintSemicolon
\KwIn{$K$, $m$, $n$, and $N$}
\KwOut{Matrix of size $K\times m$ containing $K$ initial centers}
$i \gets 1$\;
\While{$i \le N$}{
   Run K-Modes with random initialization to partition $X$ into $K$ clusters, and store the $K$ centers in mode pool $P_i$.\;
   $i \gets i + 1$\;
}
$x \gets 1$, $y \gets 1$\;
\While{$x \le K$}{
   \While{$y \le m$}{
      Extract the most frequent mode (from all $P_i$) for coordinate $(x,y)$ and assign it to the initial centers matrix $I_{x,y}$.\;
      $x \gets x + 1$, $y \gets y + 1$\;
   }
}
\caption{K-Modes initialization using Evidence Accumulation \cite{khan2007computation}.}
\label{algo:khanAndKant2007}
\end{algorithm}

\subsection{Algorithms using density and distance}
Many approaches select initial centers by computing a density measure (often based on the frequency of attribute values or distances) and then choosing centers that are both high-density and well separated.

\subsubsection{Wu et al., 2007}
Wu et al. \cite{wu2007new} proposed a density-based initial center selection method using sub-sampling and refinement. In each sub-sample, densities are computed as
\[
Dens(x_i) = \frac{1}{\sum_{j=1}^{n} d(x_i,x_j)},
\]
and the highest-density point is selected as the first center. Probabilities are computed for the remaining points, and the final centers are chosen based on minimizing the sum of square errors (SSE). See Algorithm~\ref{algo:wu2007}.

\begin{algorithm}
\DontPrintSemicolon
\KwIn{$X$, $K$, $n$}
\KwOut{$K$ initial centers}
$FCS \gets \emptyset$\;
Select $c$ sub-samples $sp_i$ from $X$ (with $c$ equal to the square of the dataset size).\;
\ForEach{$sp_i$}{
   $CS \gets \emptyset$, $k \gets 1$\;
   Compute densities for all $x \in sp_i$:
   \[
   Dens(x) = \frac{1}{\sum_{j=1}^{n} d(x,x_j)}
   \]
   Choose the point with the highest density as the first center $v_k$, and set $k\gets k+1$.\;
   Compute probabilities for remaining points (per \cite{wu2007new}) and store them in $P$.\;
   \While{$k \le c$}{
      Choose the point with maximum probability in $P$ as $v_k$, remove it from $P$, and set $k\gets k+1$.\;
   }
   Compute new cluster centers $cs_i$ by performing one iteration of clustering using $v_k$ as initial centers; let $CS$ be the union of these.\;
   Cluster $CS$ $c$ times (using $cs_i$ as initial centers) to obtain final centers $fcs_i$, and let $FCS$ be their union.\;
   Compute the SSE:
   \[
   SSE = \sum_{i=1}^{n} d(x_i,v_{c(i)}),
   \]
   where $v_{c(i)}$ is the center to which $x_i$ belongs. Choose the set corresponding to the minimum SSE.
}
\caption{K-Modes initialization using sub-sampling \cite{wu2007new}}
\label{algo:wu2007}
\end{algorithm}

\subsubsection{Cao, Liang, and Bai, 2009}
Cao et al. \cite{cao2009new} proposed an initialization method that computes density based on attribute frequencies and selects centers by combining density and distance. The algorithm is summarized in Algorithm~\ref{algo:Cao2009}.

\begin{algorithm}
\DontPrintSemicolon
\KwIn{$X$, $M$, $K$, $m$ \quad {\bf Output:} $K$ Centers}
$Centers \gets \emptyset$\;
\ForEach{$x_i \in X$}{
   Calculate 
   \[
   Dens(x_i)=\frac{1}{m}\sum_{a \in M} Dens_a(x_i),
   \]
   where $Dens_a(x_i)$ is the number of matches for attribute $a$ between $x_i$ and all $y \in X$.
}
Select the point with maximum density as the first center and add it to $Centers$.\;
Add to $Centers$ the point $x_{i_2}$ that maximizes 
\[
d(x_{i_2}, x_m)\times Dens(x_{i_2}) \quad \text{for } x_m \in Centers.
\]
If $|Centers| < K$, repeat until $K$ centers are selected.
\caption{Algorithm for K-Modes initialization using average density and distance \cite{cao2009new}.} 
\label{algo:Cao2009}
\end{algorithm}

\subsubsection{Bai et al. 2012}
Bai et al. \cite{bai2012cluster} proposed an initialization algorithm similar in spirit to the previous methods. For the first cluster, the algorithm considers the distance between an object and the overall dataset's mode along with density. The algorithm (now converted entirely to algorithm2e style) is given in Algorithm~\ref{algo:Bai2012}.

\begin{algorithm}
\DontPrintSemicolon
\KwIn{$X$, $M$, $K$, $m$}
\KwOut{$K$ Centers}
$Centers \gets \emptyset$\;
Compute densities for all $x\in X$:
\[
Dens(x) = -\frac{1}{n}\sum_{y \in X} d(x,y)
\]
Compute the mode $z$ of $X$.\;
\For{$k=1$ \KwTo $K$}{
  \eIf{$k=1$}{
    Set $PosEx(x)=Dens(x)+d(x,z)$ for all $x\in X$.
  }{
    Set $PosEx(x)=Dens(x)+\min_{v\in Centers}d(x,v)$ for all $x\in X$.
  }
  Choose $x_k = \arg\max_{x\in X} PosEx(x)$ and add it to $Centers$.\;
  Define 
  \[
  CA_{x_k}=\{ca_1,\ldots, ca_m\},
  \]
  where for $1\le j\le m$, $ca_j$ is the mode of $\{y\in X \mid d(y,x_k)\le j\}$.\;
  \For{$j=1$ \KwTo $m$}{
    Compute $Dens(ca_j)$.
  }
  \eIf{$k=1$}{
    For each $y\in CA_{x_k}$, set $PosCenter(y)=Dens(y)+d(y,z)-d(y,x_k)$.
  }{
    For each $y\in CA_{x_k}$, set $PosCenter(y)=Dens(y)+\min_{v\in Centers}d(y,v)-d(y,x_k)$.
  }
  Choose $v_k = \arg\max_{y\in CA_{x_k}} PosCenter(y)$ and update $Centers\gets Centers\cup\{v_k\}$.\;
}
\caption{Algorithm for K-Modes initialization using method \cite{bai2012cluster}.}
\label{algo:Bai2012}
\end{algorithm}

\subsection{Khan and Ahmad, 2013}
Khan and Ahmad \cite{khan2013cluster} proposed an algorithm that clusters the data based on different attribute values and yields deterministic modes for initial centers. Each data point is assigned a clustering string (e.g., $(S_{i1},S_{i2},\ldots,S_{im})$), and distinct strings are used to determine the centers. The method is summarized in Algorithm~\ref{algo:KhanAndAhmad2013}.

\begin{algorithm}
\DontPrintSemicolon
\KwIn{$X$, $n$, $M$ (attributes chosen using either all/prominent/significant), $m$, $K$, $m_i$ (number of categories for attribute $i$)}
\KwOut{$K$ Centers}
\For{$j=1$ \KwTo $m$}{
   Partition $X$ into $m_j$ clusters based on the $j$th attribute and compute their centers.\;
   Run K-Modes using these centers as initial centers and record each data point’s cluster label $S_{ij}$.\;
}
Each data point now has a clustering string $(S_{i1},S_{i2},\ldots,S_{im})$.\;
Find distinct clustering strings, count their frequencies, and sort them in descending order. Let the number of distinct strings be $K'$.\;
\If{$K' = K$}{
   Compute the modes of these $K'$ clusters and use them as the initial centers.\;
}
\ElseIf{$K' > K$}{
   Merge similar clustering strings into $K$ clusters and compute their modes.\;
}
\Else{
   Reduce $K$ and repeat the process.\;
}
\caption{Method for K-Modes initialization by clustering data multiple times based on attribute values.}
\label{algo:KhanAndAhmad2013}
\end{algorithm}

\subsection{Jiang et al. 2016}
Jiang et al. \cite{jiang2016initialization} proposed using outlier detection techniques to select initial centers for K-Modes. Their algorithms (\emph{Ini-distance} and \emph{Ini-entropy}) compute an outlier factor (based on distance or partition entropy) for each data point and then select centers based on the outlier factor and the distance from already chosen centers. The procedure is summarized in Algorithm~\ref{algo:jiang2016}.

\begin{algorithm}
\DontPrintSemicolon
\KwIn{$X$, $n$, $K$, $m$, $M$ (the set of attributes)}
\KwOut{Set $C$ of initial centers}
$C \gets \emptyset$\;
Calculate the outlier factor for all $x\in X$ (using either distance or entropy methods).\;
Select the point with the minimum outlier factor as the first center and add it to $C$.\;
\While{$|C| < K$}{
  \ForEach{$x\in X\setminus C$}{
    \For{$j=1$ \KwTo $|C|$}{
       Compute the weighted matching distance:
       \[
       wd(x,v_j)=\sum_{a\in M}\text{weight}(a)\times \delta(x,y),
       \]
       where $\delta(x,y)$ is defined in (2).
    }
    Compute the possibility of $x$ being a center based on the outlier factor and the distances.\;
  }
  Select the data point with the maximum possibility and add it to $C$.\;
}
\caption{Algorithm for K-Modes initialization using outlier detection techniques \cite{jiang2016initialization}.}
\label{algo:jiang2016}
\end{algorithm}

\subsection{Summary}
Many methods for selecting initial centers for K-Modes have been proposed. Some require setting multiple parameters or suffer from high computational costs. Among these, the method of Cao et al. \cite{cao2009new} is parameter-free and works well for large datasets, although its computation cost grows quadratically with $K$.

%%%%%%%%%%%% Section 3 %%%%%%%%%%%%
\section{The New Method for Choosing Initial Cluster Centers}
Since we target large datasets, efficiency is crucial while maintaining clustering quality. To achieve both, we employ successive bisection (as in Bisecting K-Means \cite{steinbach2000comparison,zhuang2016limited}) extended to categorical data, yielding Bisecting K-Modes. Starting with the whole dataset as a single cluster, BK-Modes iteratively partitions the dataset into two clusters, then three, and so on. Suppose at one iteration there are $J$ clusters ($J<K$); BK-Modes produces $(J+1)$ clusters by selecting one cluster (e.g., the one with the largest sum of distances) and bisecting it using the Two-Modes algorithm (i.e., K-Modes with $K=2$). This process continues until $J=K$. The final $K$ centers from BK-Modes are then used as the initial centers for K-Modes.

\RestyleAlgo{boxruled}
\LinesNumbered
\begin{algorithm}
  \DontPrintSemicolon
  \KwIn{Dataset \textbf{D}, integer $k$}
  \KwOut{$k$ clusters}\;
  $j \gets 1$, $clusters\_set \gets \{\textbf{D}\}$\;
  \While{$j < k$}{
    Select a cluster $C$ from $clusters\_set$ based on a criterion (e.g., largest sum of distances).\;
    Bisect $C$ using the Two-Modes algorithm (see Algorithm~\ref{algo:twomodes}) to obtain clusters \textbf{C1} and \textbf{C2}.\;
    Update $clusters\_set \gets (clusters\_set \setminus \{C\}) \cup \{\textbf{C1},\, \textbf{C2}\}$.\;
    $j \gets j + 1$\;
  }
  Run K-Modes initialized with the $K$ centers from $clusters\_set$ until convergence.\;
  \caption{{\sc Bisecting K-Modes}}
  \label{algo:bisecting}
\end{algorithm}

At each iteration, two decisions are made:
\begin{itemize}
\item \textbf{Cluster Selection:} Choose the cluster with the largest sum of distances.
\item \textbf{Two-Modes Initialization:} In the selected cluster, choose its mode as one initial center and the farthest point from the mode as the other.
\end{itemize}

The Two-Modes clustering algorithm is given below.

\RestyleAlgo{boxruled}
\LinesNumbered
\begin{algorithm}
  \DontPrintSemicolon
  \KwIn{Cluster \textbf{D}}
  \KwOut{Sub-clusters \textbf{D1} and \textbf{D2}}
  Set $C1 \gets$ mode of \textbf{D}\;
  Set $C2 \gets$ the farthest data point from $C1$\;
  \While{cluster assignments change}{
    \ForEach{data point $P$ in \textbf{D}}{
      \eIf{$d(P,C1) < d(P,C2)$}{assign $P$ to \textbf{D1}}{assign $P$ to \textbf{D2}}
    }
    Update $C1 \gets$ mode of \textbf{D1} and $C2 \gets$ mode of \textbf{D2}\;
  }
  \Return{\textbf{D1}, \textbf{D2}}\;
  \caption{{\sc Two-Modes Clustering Algorithm}}
  \label{algo:twomodes}
\end{algorithm}

%%%%%%%%%%%%%%%%%%%%%%%%%%%%%%%%%%%%%%%%%%%%%%%%%%%%%%%%%%%%%%%%%%%%%%%%%%%%%%%%%%%%
\section{Experimental studies}

\subsection{Datasets} %%%%%%%%%%%% Sec. 4.1 %%%%%%%%%%%%
To evaluate the performance of our method, we choose three datasets:
\begin{enumerate}
\item \textbf{US Census Dataset:} 2,458,284 observations with 68 attributes (the first attribute is dropped). The maximum number of categories is 17.
\item \textbf{KDD Cup 1999 Dataset:} 4,898,430 data points with 42 attributes. Attributes with 256 or more categories were dropped, leaving 35 attributes (Columns 1, 5, 6, 13, 14, 32, and 33 are removed). The maximum number of categories is 100.
\item \textbf{PUF Dataset:} Generated from a 64-stage 5-XOR PUF instance, with 4,995,137 challenges. Each challenge has 64 binary attributes.
\end{enumerate}

\subsection{Evaluation Metrics} %%%%%%%%%%%% Sec. 4.2 %%%%%%%%%%%% 
Clustering quality is measured by the sum of distances (SD)
\begin{equation}
SD_{total} = \frac{1}{n} \sum_{j=1}^{K} \sum_{x_i\in C_j} d(x_i,\mu_j),
\end{equation}
where $\mu_j$ is the center of cluster $C_j$. Lower SD indicates better clustering quality. Efficiency is measured by the total execution time (including initialization and K-Modes iterations) and the number of iterations.

\subsection{Experimental Set-up} %%%%%%%%%%%% Sec. 4.3 %%%%%%%%%%%%
We implemented our clustering method in C, representing categorical values using 8-bit unsigned integers. Each dataset was pre-processed by recoding attribute values as $0, 1, \ldots, m_i-1$. We also implemented the method of Cao et al. (Sec. 2.5.2) and random initialization (using 5 different sets of initial centers). Experiments were run on a single core of a node at Texas Tech University (dual Xeon E5-2695v4, 192 GB, CentOS 7.4 with OpenHPC).

\subsection{Experimental Results} %%%%%%%%%%%% Sec. 4.4 %%%%%%%%%%%%
We tested three methods for choosing initial centers for K-Modes: random initialization, the method of Cao et al., and our proposed method. For random initialization, five different sets of initial centers were used. The resulting initial centers were then fed to K-Modes, which iterated until convergence. Tables~\ref{tab:usresults}--\ref{tab:pufresults} list the results, where “SD” is the sum of distances after convergence, “Iterations” are the number of K-Modes iterations, and “Time” is the total execution time.

\begin{table*}[t]
\def\arraystretch{1.1}%
\centering
\caption{Clustering results for Dataset 1}
\label{tab:usresults}
\begin{tabular}{|c|c|l|l|l|l|}
\hline
\multicolumn{1}{|l|}{\textbf{Different values of K}} & \multicolumn{1}{l|}{\textbf{Methods}} & \textbf{K initial Centers} & \textbf{SD} & \textbf{Iterations} & \textbf{Time} \\ \hline
\multirow{7}{*}{\textbf{K=30}}             & \multirow{5}{*}{\textbf{Random}}                & Set1            & 9.45       & 10                   & 1 min 18 secs                   \\ \cline{3-6} 
                                           &                                                 & Set2            & 10.25      & 36                   & 6 min                         \\ \cline{3-6} 
                                           &                                                 & Set3            & 10.20      & 35                   & 5 min 16 sec                  \\ \cline{3-6} 
                                           &                                                 & Set4            & 9.91       & 40                   & 6 min 21 sec                 \\ \cline{3-6} 
                                           &                                                 & Set5            & 10.20      & 35                   & 5 min 23 sec                 \\ \cline{2-6} 
                                           & \multicolumn{2}{c|}{\textbf{Cao}}                               & 9.87       & 10                   & 5 min 36 sec                 \\ \cline{2-6} 
                                           & \multicolumn{2}{c|}{\textbf{Proposed}}                          & 9.45       & 5                    & 2 min                        \\ \hline
\multirow{7}{*}{\textbf{K=100}}            & \multirow{5}{*}{\textbf{Random}}                & Set1            & 9.07       & 108                  & 47 min                       \\ \cline{3-6} 
                                           &                                                 & Set2            & 8.64       & 96                   & 45 min                       \\ \cline{3-6} 
                                           &                                                 & Set3            & 9.03       & 106                  & 46 min                       \\ \cline{3-6} 
                                           &                                                 & Set4            & 8.22       & 10                   & 8 min                        \\ \cline{3-6} 
                                           &                                                 & Set5            & 9.18       & 105                  & 44 min                       \\ \cline{2-6} 
                                           & \multicolumn{2}{c|}{\textbf{Cao}}                               & 8.89       & 13                   & 58 min                       \\ \cline{2-6} 
                                           & \multicolumn{2}{c|}{\textbf{Proposed}}                          & 8.06       & 5                    & 5 min                        \\ \hline
\multirow{7}{*}{\textbf{K=300}}            & \multirow{5}{*}{\textbf{Random}}                & Set1            & 8.14       & 310                  & 7 hr 35 min                  \\ \cline{3-6} 
                                           &                                                 & Set2            & 7.81       & 286                  & 7 hr 47 min                  \\ \cline{3-6} 
                                           &                                                 & Set3            & 8.12       & 297                  & 7 hr 19 min                  \\ \cline{3-6} 
                                           &                                                 & Set4            & 7.22       & 11                   & 24 min                       \\ \cline{3-6} 
                                           &                                                 & Set5            & 8.19       & 308                  & 7 hr 39 min                  \\ \cline{2-6} 
                                           & \multicolumn{2}{c|}{\textbf{Cao}}                               & 8.05       & 12                   & 7 hr 29 min                  \\ \cline{2-6} 
                                           & \multicolumn{2}{c|}{\textbf{Proposed}}                          & 7.07       & 8                    & 20 min                       \\ \hline
\end{tabular}
\vspace{3mm}
\end{table*}

\begin{table*}[t]
\def\arraystretch{1.1}%
\caption{Clustering results for Dataset 2}
\label{tab:kddresults}
\centering
\begin{tabular}{|c|c|l|l|l|l|}
\hline
\multicolumn{1}{|l|}{\textbf{Different values of K}} & \multicolumn{1}{l|}{\textbf{Methods}} & \textbf{K Initial Centers} & \textbf{SD} & \textbf{Iterations} & \textbf{Time} \\ \hline
\multirow{7}{*}{\textbf{K=30}}             & \multirow{5}{*}{\textbf{Random}}                & Set1            & 1.05       & 33                   & 5 min 50 sec                 \\ \cline{3-6} 
                                           &                                                 & Set2            & 1.05       & 20                   & 4 min 47 sec                 \\ \cline{3-6} 
                                           &                                                 & Set3            & 1.05       & 34                   & 5 min 54 sec                 \\ \cline{3-6} 
                                           &                                                 & Set4            & 1.06       & 37                   & 6 min 25 sec                 \\ \cline{3-6} 
                                           &                                                 & Set5            & 1.05       & 34                   & 6 min 38 sec                 \\ \cline{2-6} 
                                           & \multicolumn{2}{c|}{\textbf{Cao}}                               & 0.90       & 8                    & 5 min 43 sec                 \\ \cline{2-6} 
                                           & \multicolumn{2}{c|}{\textbf{Proposed}}                          & 0.82       & 7                    & 2 min 40 sec                 \\ \hline
\multirow{7}{*}{\textbf{K=100}}            & \multirow{5}{*}{\textbf{Random}}                & Set1            & 0.99       & 103                  & 49 sec                       \\ \cline{3-6} 
                                           &                                                 & Set2            & 0.97       & 107                  & 56 min                       \\ \cline{3-6} 
                                           &                                                 & Set3            & 1.06       & 64                   & 22 min                       \\ \cline{3-6} 
                                           &                                                 & Set4            & 0.99       & 84                   & 31 sec                       \\ \cline{3-6} 
                                           &                                                 & Set5            & 1.01       & 50                   & 24 min                       \\ \cline{2-6} 
                                           & \multicolumn{2}{c|}{\textbf{Cao}}                               & 0.83       & 12                   & 1 hr 5 min                   \\ \cline{2-6} 
                                           & \multicolumn{2}{c|}{\textbf{Proposed}}                          & 0.61       & 11                   & 10 min                       \\ \hline
\multirow{7}{*}{\textbf{K=300}}            & \multirow{5}{*}{\textbf{Random}}                & Set1            & 0.84       & 127                  & 2 hr 39 min                  \\ \cline{3-6} 
                                           &                                                 & Set2            & 0.79       & 174                  & 3 hr 15 min                  \\ \cline{3-6} 
                                           &                                                 & Set3            & 1.05       & 64                   & 36 min                       \\ \cline{3-6} 
                                           &                                                 & Set4            & 0.99       & 84                   & 51 min                       \\ \cline{3-6} 
                                           &                                                 & Set5            & 0.83       & 187                  & 3 hr 40 min                  \\ \cline{2-6} 
                                           & \multicolumn{2}{c|}{\textbf{Cao}}                               & 0.68       & 10                   & 9 hr 3 min                   \\ \cline{2-6} 
                                           & \multicolumn{2}{c|}{\textbf{Proposed}}                          & 0.45       & 9                    & 23 min                       \\ \hline
\end{tabular}
\end{table*}

\begin{table*}[t]
\def\arraystretch{1.1}%
\caption{Clustering results for Dataset 3}
\label{tab:pufresults}
\centering
\begin{tabular}{|c|c|l|l|l|l|}
\hline
\multicolumn{1}{|l|}{\textbf{Different values of K}} & \multicolumn{1}{l|}{\textbf{Methods}} & \textbf{K Initial Centers} & \textbf{SD} & \textbf{Iterations} & \textbf{Time} \\ \hline
\multirow{7}{*}{\textbf{K=30}}          & \multirow{5}{*}{\textbf{Random}}      & Set1            & 24.10      & 2                    & 56 sec                     \\ \cline{3-6} 
                                        &                                       & Set2            & 23.70      & 28                   & 8 min 7 sec                \\ \cline{3-6} 
                                        &                                       & Set3            & 23.68      & 30                   & 8 min 34 sec               \\ \cline{3-6} 
                                        &                                       & Set4            & 23.70      & 31                   & 8 min 20 sec               \\ \cline{3-6} 
                                        &                                       & Set5            & 23.68      & 30                   & 8 min 20 sec               \\ \cline{2-6} 
                                        & \multicolumn{2}{c|}{\textbf{Cao}}      & 24.60      & 2                    & 7 min 31 sec               \\ \cline{2-6} 
                                        & \multicolumn{2}{c|}{\textbf{Proposed}} & 23.85      & 2                    & 1 min 48 sec               \\ \hline
\multirow{7}{*}{\textbf{K=100}}         & \multirow{5}{*}{\textbf{Random}}      & Set1            & 21.93      & 99                   & 2 hr 5 min                 \\ \cline{3-6} 
                                        &                                       & Set2            & 21.94      & 91                   & 1 hr 18 min                \\ \cline{3-6} 
                                        &                                       & Set3            & 22.44      & 2                    & 3 min                      \\ \cline{3-6} 
                                        &                                       & Set4            & 21.93      & 101                  & 2 hr 3 min                 \\ \cline{3-6} 
                                        &                                       & Set5            & 21.93      & 100                  & 2 hr 2 min                 \\ \cline{2-6} 
                                        & \multicolumn{2}{c|}{\textbf{Cao}}      & 22.80      & 2                    & 1 hr 15 min                \\ \cline{2-6} 
                                        & \multicolumn{2}{c|}{\textbf{Proposed}} & 22.07      & 2                    & 5 min                      \\ \hline
\multirow{7}{*}{\textbf{K=300}}         & \multirow{5}{*}{\textbf{Random}}      & Set1            & 20.53      & 271                  & 14 hr 49 min               \\ \cline{3-6} 
                                        &                                       & Set2            & 21.07      & 2                    & 10 min                     \\ \cline{3-6} 
                                        &                                       & Set3            & 20.53      & 295                  & 15 hr 28 min               \\ \cline{3-6} 
                                        &                                       & Set4            & 20.53      & 301                  & 15 hr 24 min               \\ \cline{3-6} 
                                        &                                       & Set5            & 20.53      & 300                  & 15 hr 23 min               \\ \cline{2-6} 
                                        & \multicolumn{2}{c|}{\textbf{Cao}}      & 21.35      & 2                    & 15 hr 5 min                \\ \cline{2-6} 
                                        & \multicolumn{2}{c|}{\textbf{Proposed}} & 20.65      & 2                    & 11 min                     \\ \hline
\end{tabular}
\end{table*}

Figures~1--3 (below) show plots comparing SD and computation time for the three methods.

\begin{figure*}[t]
  \centering
  \subfloat[Sum of distances measure for each method]{\includegraphics[width=3.1in]{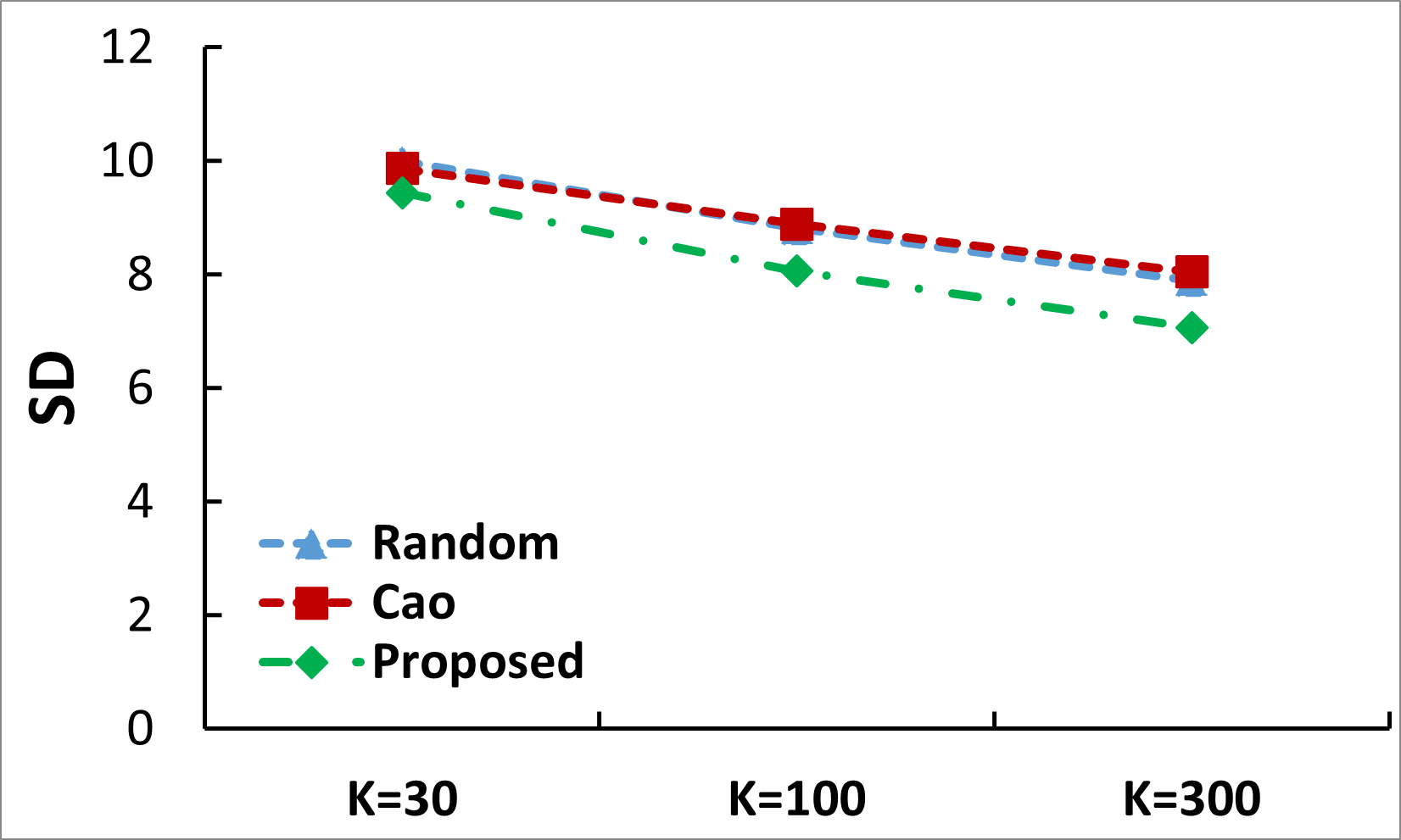}\label{graph11_1}}
  \hfill
  \subfloat[Computation time measure for each method]{\includegraphics[width=3.1in]{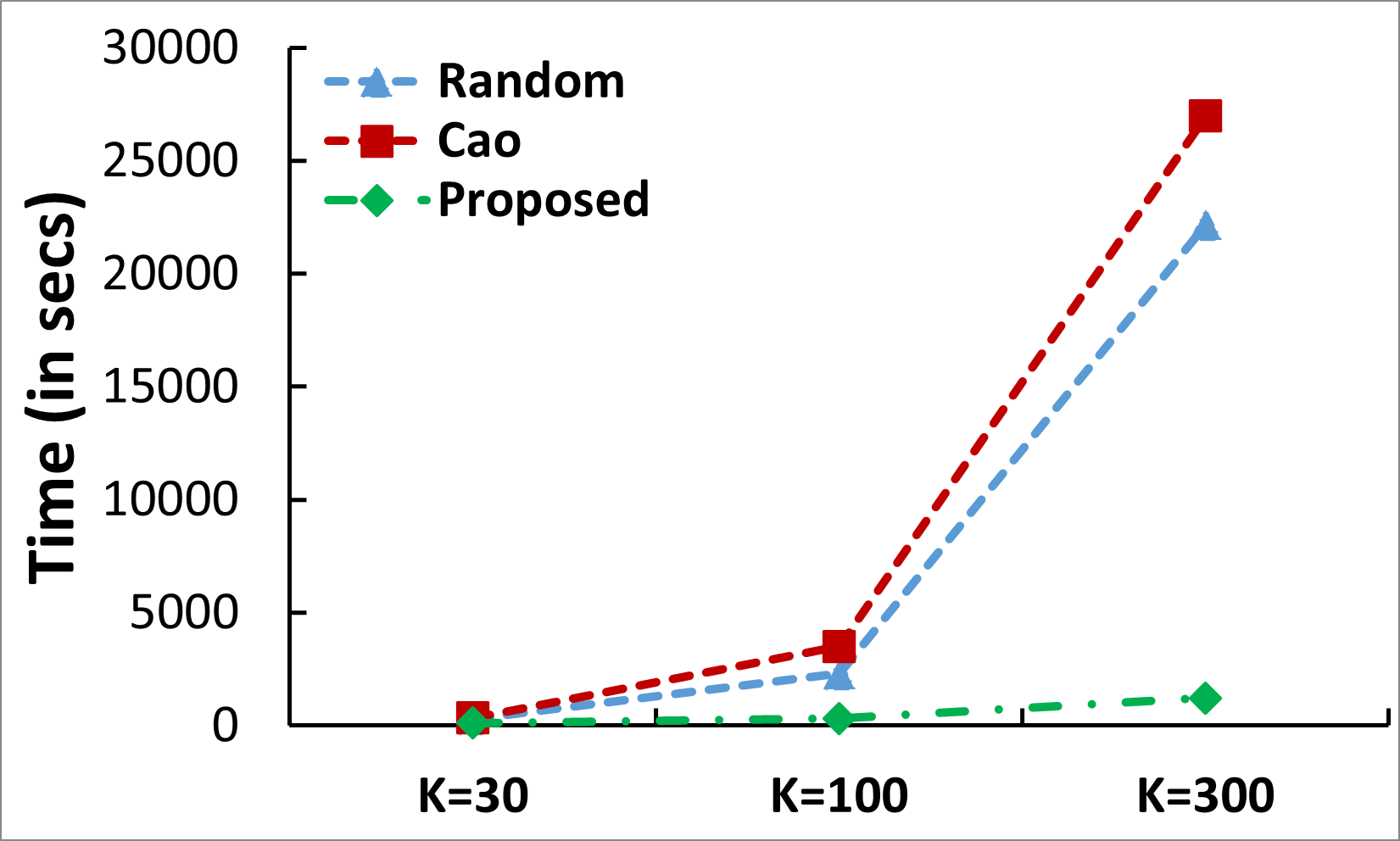}\label{graph12_1}}
  \caption{Results of three methods on Dataset 1.}
  \label{fig:Accuracy_Compersion1}
\end{figure*}

\vspace{3mm}

\begin{figure*}[t]
  \centering
  \subfloat[Sum of distances measure for each method]{\includegraphics[width=3.1in]{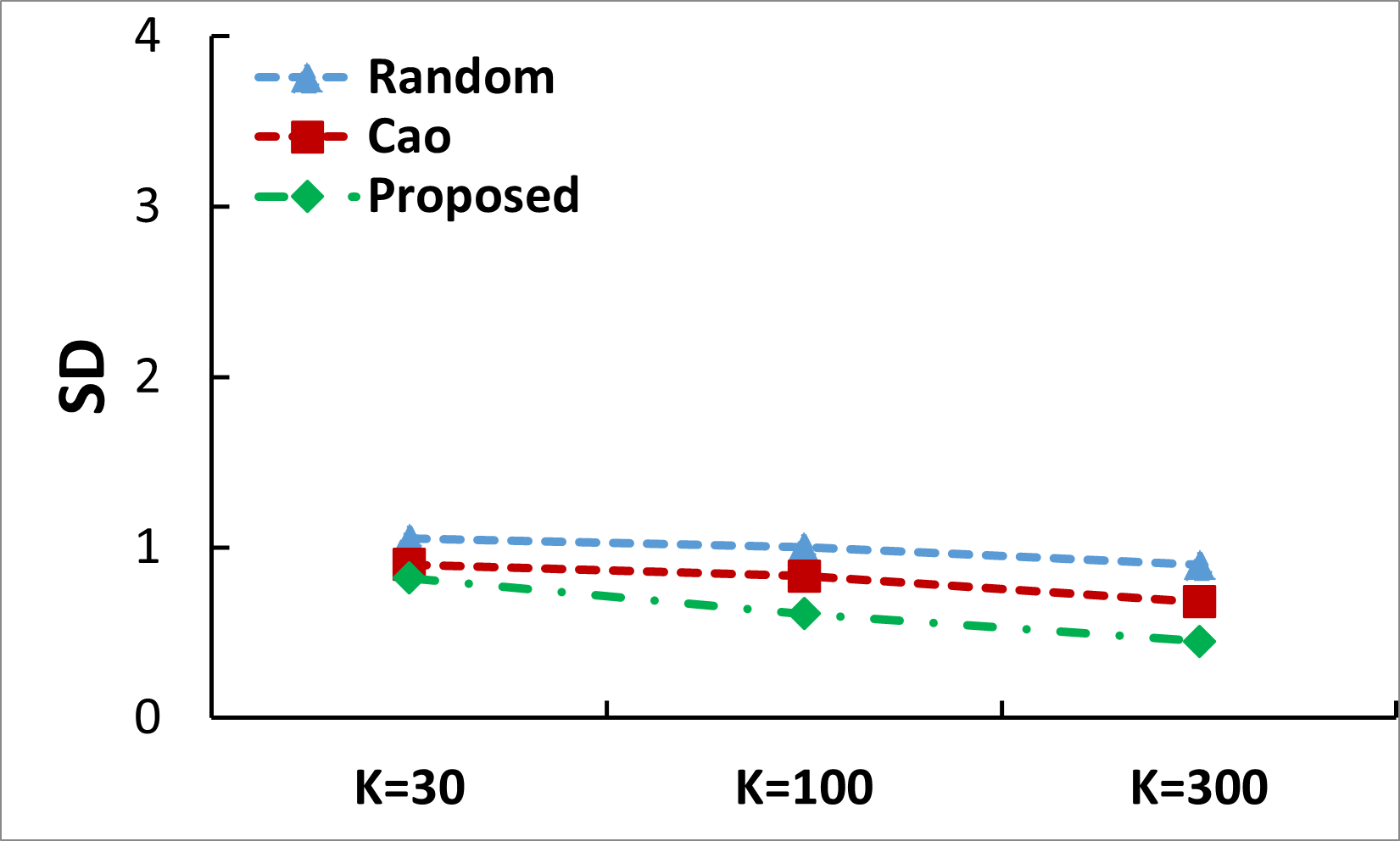}\label{graph31_1}}
  \hfill
  \subfloat[Computation time measure for each method]{\includegraphics[width=3.1in]{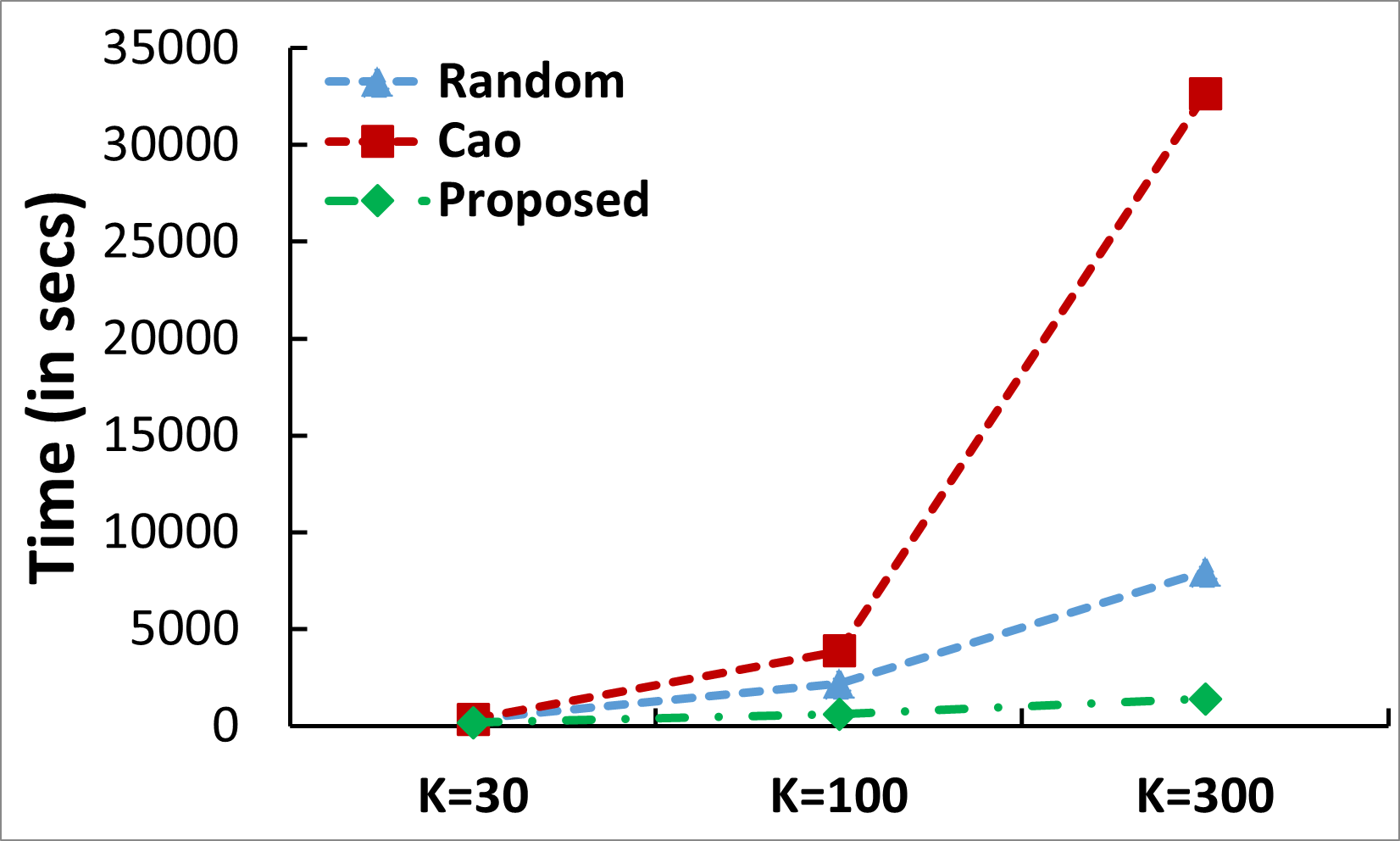}\label{graph32_1}}
  \caption{Results of three methods on Dataset 2.}
  \label{fig:Accuracy_Compersion2}
\end{figure*}

\begin{figure*}[t]
  \centering
  \subfloat[Sum of distances measure for each method]{\includegraphics[width=3.1in]{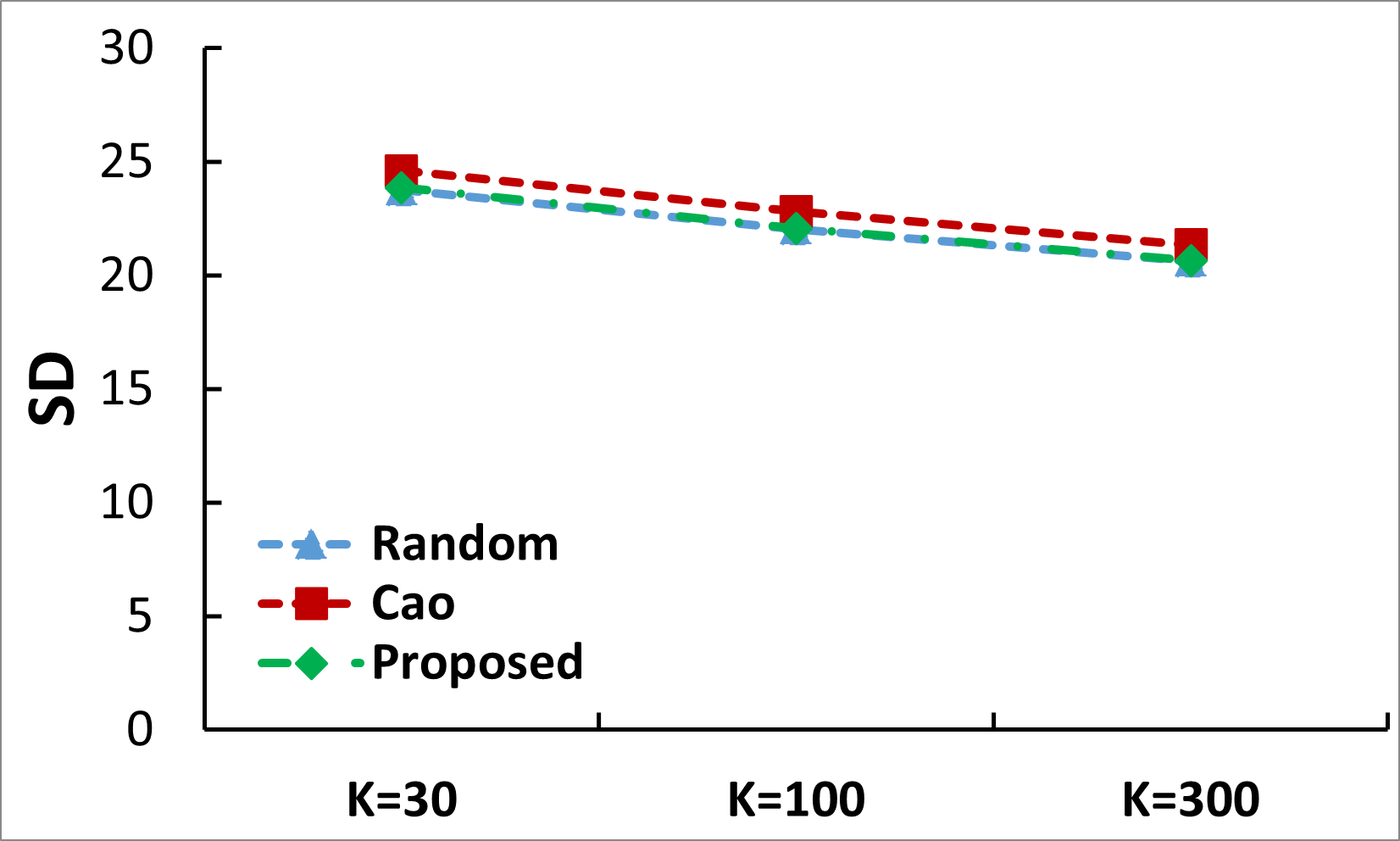}\label{graph41_1}}
  \hfill
  \subfloat[Computation time measure for each method]{\includegraphics[width=3.1in]{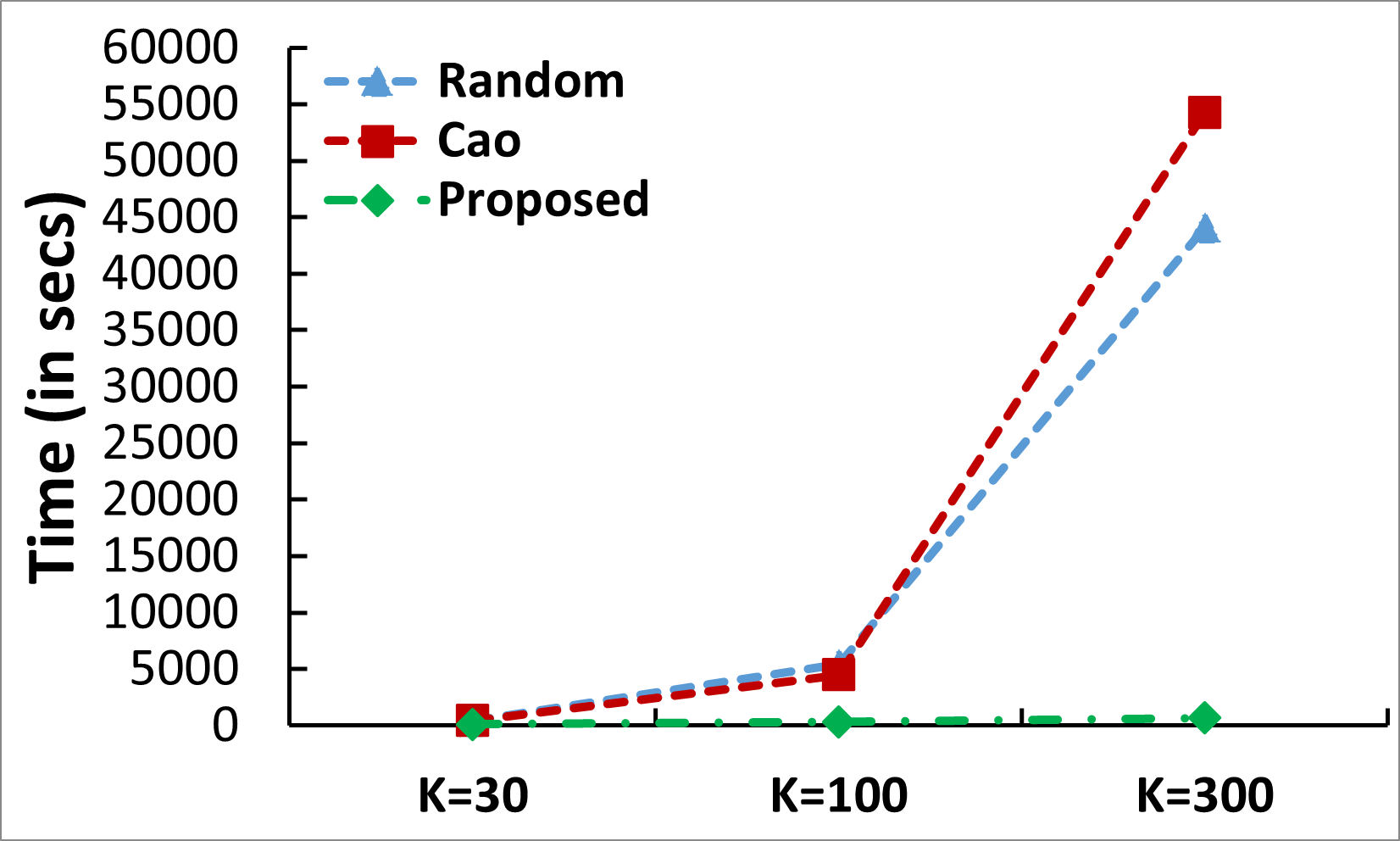}\label{graph42_1}}
  \caption{Results of three methods on Dataset 3.}
  \label{fig:Accuracy_Compersion3}
\end{figure*}

\section{Conclusion}
Big datasets with many attributes are computationally challenging to analyze, and clustering is a popular approach to partition such data into smaller, more homogeneous subsets that are easier to analyze. For categorical data, the K-Modes clustering algorithm is the method of choice; however, it requires $K$ initial centers, and random initialization may lead to poor clustering quality and a large number of iterations. Thus, it is important to find methods for selecting initial centers that are both efficient and produce high-quality clusters. While several effective methods exist, many require the setting of multiple parameters or suffer from increased computation as $K$ increases.

We propose Bisecting K-Modes, which iteratively partitions the dataset using the Two-Modes algorithm (i.e., K-Modes with $K=2$) until $K$ clusters are obtained. Experimental tests on several large categorical datasets show that Bisecting K-Modes is both efficient and produces high clustering quality, making it a reliable high-performance method.

\section{Acknowledgements}
The research was supported in part by the National Science Foundation under Grant No. CNS-1526055 and Grant No. OAC-2103563.

\bibliography{bib} 
\bibliographystyle{unsrtnat}

\end{document}